# Fast and Optimal Laplacian Solver for Gradient-Domain Image Editing using Green Function Convolution

Dominique Beaini, Sofiane Achiche, Fabrice Nonez, Olivier Brochu Dufour, Cédric Leblond-Ménard, Mahdis Asaadi, Maxime Raison

Abstract
In computer vision, the gradient and Laplacian of an image are used in different applications, such as edge detection, feature extraction, and seamless image cloning. Computing the gradient of an image is straightforward since numerical derivatives are available in most computer vision toolboxes. However, the reverse problem is more difficult, since computing an image from its gradient requires to solve the Laplacian equation (also called Poisson equation). Current discrete methods are either slow or require heavy parallel computing. The objective of this paper is to present a novel fast and robust method of solving the image gradient or Laplacian with minimal error, which can be used for gradient-domain editing. By using a single convolution based on a numerical Green's function, the whole process is faster and straightforward to implement with different computer vision libraries. It can also be optimized on a GPU using fast Fourier transforms and can easily be generalized for an n-dimension image. The tests show that, for images of resolution 801x1200, the proposed GFC can solve 100 Laplacian in parallel in around 1.0 milliseconds (ms). This is orders of magnitude faster than our nearest competitor which requires 294ms for a single image. Furthermore, we prove mathematically and demonstrate empirically that the proposed method is the least-error solver for gradient domain editing. The developed method is also validated with examples of Poisson blending, gradient removal, and the proposed gradient domain merging (GDM). Finally, we present how the GDM can be leveraged in future works for convolutional neural networks (CNN).

**Keywords**  Computer vision · Poisson image editing · seamless cloning · Green function convolution · Gradient Laplacian Solver · gradient domain editing

## 1 Introduction

In computer vision and signal processing, the images can be interpreted as numerical potentials, especially when there is an interest in their gradient ($E$). For example, early computer vision system algorithms relied a lot on numerical gradients and Laplacian [1–6] to extract important information about edges and image boundaries. They are computed using simple convolution kernel such as Sobel [7–9]. More recently, there have been growing interest in gradient domain image editing (GDIE) applications, which aim at editing or creating images from its gradient [1–3, 10, 11].

Although computing the gradient or Laplacian of an image is straightforward, the reverse problem of computing the image from its gradient is a non-trivial task. In fact, this problem requires to solve a differential equation [1–3] without knowing if an exact solution exists. When the gradient is computed from an image, it always generates a conservative field, meaning that the field can be integrated to obtain a potential (the original image). For gradient domain image editing (GDIE), a non-conservative perturbation is voluntarily introduced to the gradient, meaning that the resulting field cannot be integrated into an exact solution.

Nevertheless, it is still interesting to solve the non-conservative gradient since it leads to many gradient-domain editing applications, such as gradient erasing, seamless cloning and vectorization with diffusion curves [1–3, 10, 11]. Furthermore, Bhat et al. presented a whole framework of gradient-domain image editing with unique and useful applications such as color filtering and edge sharpening [12].

The first method to solve the image Laplacian (also called Poisson's equation) was presented in 2003 by Perez et al. [2], which proposed to solve the differential equation by iteratively minimizing the variational problem. Other research followed by optimizing the computation speed and error [2, 10], others used the Jacobi method [1], and McCann proposes a multi-grid solver [11]. These approaches converge to the approximate solution, but they are harder to implement since they are iterative, which also makes them slower to compute. An alternate way of solving the Poisson equation is proposed by Tanaka [3] by modifying the Poisson problem into a closed-form problem using cosine transforms.

More recent methods solve this problem by using methods based on multipoles and Green's function [13–15]. However, they are only implemented for diffusion curves in vector graphics, and not for gradient domain image editing. Furthermore, the Green's function methods [13–15] and the Tanaka method [3] propose analytical solutions for continuous space, but we propose a numerical solution developed for a discrete space. Hence, our method is more suitable for discrete images and is demonstrated to have a lower error than Tanaka in section 3.1.4.

The objective of this paper is to present a novel fast and robust method of solving the image gradient or Laplacian with minimal error, which can be used for gradient-domain editing.

In the research work presented here, a novel method is proposed called Green Function Convolution (GFC), which allows solving any modified gradient. In the case of a non-conservative field, the proposed GFC method is proven to find the best possible approximation in terms of gradient error. In fact, we mathematically prove in section 2.2 and empirically in section 3.1.4 that GFC is the optimal possible solver for any perturbation added to the gradient, meaning that gradient domain editing can be done with minimal error.

Our contributions are summarized below:

**Simple, fast and optimal gradient/Laplacian solver.** The implementation that we propose is simple, requiring only a few lines of code using any library that implements the 2D fast Fourier transform (FFT). The implementation is also significantly faster than competing methods since we showed in **Fig. 2** a 170x improvement compared to Tanaka's method [3], thanks to our graphics processing units (GPU) implementation. We also showed that our GFC solver can process 100 images in 1ms using Pytorch, making it the fastest method available for discrete images. Finally, we demonstrated mathematically in section 2.2 and empirically in section 3.1.4 that GFC is optimal in the sense that it is the least-error solver for gradient domain editing.

**Gradient domain merging applications.** Inspired by edge saliency sharpening techniques [12] and recent edge detection methods [16, 17], we develop a novel method of reducing texture information and enhancing boundaries contrast. Our work proposes the first method to use machine learning edge detectors for this purpose. With our GFC solver that relies mainly on FFT, we show that the solver can be implemented in deep learning libraries such as Tensorflow and Pytorch and can be leveraged in future works for machine learning applications.

# 2 Computing the image from its gradient or Laplacian

To understand how to compute the image from its gradient field or Laplacian, we first focus on the mathematical understanding of the Green's function and its ability to solve any Laplacian [18]. We will show how to find the appropriate Green's function and how to solve either the gradient or the Laplacian. Then, we will demonstrate mathematically that using Green's function is the optimal tool when there is a non-conservative perturbation that is added to the gradient field.

## 2.1 Green's function to solve the Laplacian

This subsection explains how the Green's function can be used to theoretically solve a Laplacian

(Poisson equation) on any signal.

First, we define the gradient field $\boldsymbol{E}$ of a signal (image) $I$ in equation (1), where $\nabla$ is the $n^{th}$ dimension gradient operator. In many applications such as computer vision, computing the gradient is very simple to do using numerical derivatives such as the Sobel method [7, 19]. However, the reverse problem of finding the signal (or image) $I$ from the field $\boldsymbol{E}$ defined in (2) is not trivial, since the curvilinear integral $\int_C$ is not always defined. In fact, the integral (2) is only defined in the case of a conservative field. In the case of gradient domain image editing (GDIE), the field is modified via a non-conservative perturbation, which renders equation (2) unsolvable.

$$\boldsymbol{E} = \nabla I \quad (1)$$

$$V_E = -\int_C \boldsymbol{E} \cdot dl \quad (2)$$

Instead of solving the gradient, most approaches focus on solving the Laplacian (also known as Poisson equation) defined in (3).

$$L = \nabla \cdot \boldsymbol{E} = \nabla^2 I \quad (3)$$

Since the Laplacian is a differential equation, we propose to solve it using a Green's function, which is defined as a function that solves a differential equation via convolution [18]. This definition is expressed in (4), where $V_{mono}$ is the Green's function of the Laplacian $\nabla^2$, $\nabla \cdot$ is the divergence operator and $*$ is the convolution operator. The notation $V_{mono}$ is chosen since it is based on our previous work concerning electromagnetic potentials in images [4, 5], where the potentials are in fact the 2D Green's function [18]. The equation (4) is at the heart of our proposed Green function convolution (GFC) method.

$$\begin{aligned} I &= (\nabla^2 I) * V_{mono} \\ I &= (\nabla \cdot \boldsymbol{E}) * V_{mono} \end{aligned} \quad (4)$$

Other GDIE methods proposed using multipoles and Green's function based solvers [13, 15]. However, we differentiate ourselves from their work [13, 15] by focusing on a purely numerical solution, instead of solving the Green's function analytically.

The Green's function $V_{mono}$ is given in equation (5), with the constant $S_{n-1}$ given in equation (6) where $\Gamma$ is the gamma function and $r$ is the Euclidean distance [4, 13, 15, 18]. For the other methods based on the Green's function [13–15], $V_{mono}$ is modified to account for the rectangular boundary around the image, which is not required for us since we compute $V_{mono}$ numerically.

$$V_{mono} = \frac{-1}{S_{n-1}} \int r^{(1-n)} dr$$

$$V_{mono} = \frac{-1}{S_{n-1}} \cdot \begin{cases} ln(r), & n = 2 \\ \dfrac{r^{2-n}}{n-2}, & n \neq 2 \end{cases}, \quad n \in \mathbb{N}^* \quad (5)$$

$$S_{n-1} = \frac{2\pi^{n/2}}{\Gamma(n/2)}, \qquad S_{n=2} = 2\pi \quad (6)$$

In our previous work [4, 5], we used a physics-inspired method, which convolved electromagnetic dipoles in the direction of the gradient for partial contour analysis. Those dipole potentials $V_{dip}$ are in fact the Green's function of the gradient $\boldsymbol{E}$, meaning that they directly solve the gradient without first computing the Laplacian. The gradient solver using $V_{dip}$ is presented in equation (7), where $n$ is the number of dimensions ($n = 2$ for an image) and $x_i$ is the axis of each dimension. Hence, each dipole is convolved with each component of the gradient. Notice that the definition consists of moving the divergence operator $\nabla \cdot$ from $\boldsymbol{E}$ to $V_{mono}$ in equation (4).

$$I = \sum_{i=1}^{n} E_{x_i} * \underbrace{\frac{\partial}{\partial x_i}(V_{mono})}_{\equiv V_{dip}^i} \quad (7)$$

Although definitions (4) and (7) are both valid, the current paper focuses on the definition given by (4) since it requires a single convolution. Furthermore, $V_{dip}$ was developed in previous work to account for the strong electromagnetic inspiration. However, since it is no longer important in our current work, we will favor solving the gradient and Laplacian using equation (4).

## 2.2 Proof of optimal result for any perturbations in the gradient

The above-presented mathematical equations (4) and (7) showed how to re-compute the image $I$ from its gradient $\boldsymbol{E}$ or Laplacian $L$ using the convolutions with the Green's function $V_{mono}$. However, there are many GDIE applications that require adding a voluntary non-conservative perturbation to the field, such as those presented in section 3.2. The perturbed field is noted $\boldsymbol{E}_p$, while the computed field and potentials are respectively $\boldsymbol{E}_c$ and $I_c$.

Since the perturbation can be non-conservative, the field $\boldsymbol{E}_p$ does not have an associated potential and cannot be solved exactly. Hence, there is a need to find the conservative field $\boldsymbol{E}_c$ that is the best possible approximation of $\boldsymbol{E}_p$. This section will prove that equations (4) and (7) give the optimal $I_c$ and $\boldsymbol{E}_c$ for any possible perturbation. Thus, it proves that the proposed GFC method is robust to perturbation and that it will converge to the least error solution, where the error is defined as $\epsilon = |\boldsymbol{E}_p - \boldsymbol{E}_c|$.

First, using Hilbert projection theorem, we know that the minimum-error solution is given when $\epsilon$ is orthogonal to any conservative field $\nabla U$ at any point [20]. Hence, we need to prove that $F = 0$ (equation (8)), where $d\mu$ is the infinitesimal hyper-volume for the integration.

$$F = \int_{\mathbb{R}^n} \left[ \underbrace{[\boldsymbol{E}_p - \boldsymbol{E}_c]}_{\equiv \epsilon} \cdot \nabla U \right] d\mu = 0 \quad (8)$$

To prove (8), we first replace the value of $\boldsymbol{E}_c$ by its correspondence $\boldsymbol{E}_p$, as given in equation (9).

Then, we substitute $I_c$ by $\left((\nabla \cdot \boldsymbol{E}_p) * V_{mono}\right)$ according to equation (4). We also define the variable $A$ as a temporary variable to make it easier to follow the proof.

$$F = \int_{\mathbb{R}^n} \left[ [\boldsymbol{E}_p - \nabla(I_c)] \cdot \nabla U \right] d\mu$$

$$F = \int_{\mathbb{R}^n} \left[ \underbrace{\left[\boldsymbol{E}_p - \nabla \left((\nabla \cdot \boldsymbol{E}_p) * V_{mono}\right)\right]}_{\equiv A} \cdot \nabla U \right] d\mu \quad (9)$$

$$F = \int_{\mathbb{R}^n} [A \cdot \nabla U] \, d\mu$$

By adding and subtracting the term $(\nabla \cdot A)U$ inside the integral, we obtain equation (10). Then, we use the divergence properties in equation (11) to regroup the positive terms inside an integral and the negative terms in another.

$$F = \int_{\mathbb{R}^n} [(A \cdot \nabla U) + (\nabla \cdot A)U - (\nabla \cdot A)U] d\mu \quad (10)$$

$$F = \underbrace{\int_{\mathbb{R}^n} [\nabla \cdot (AU)] \, d\mu}_{\equiv B} - \int_{\mathbb{R}^n} (\nabla \cdot A) U d\mu \quad (11)$$

In equation (11), the term noted $B$ has a value of 0 and is canceled. This is due to Gauss's theorem which states that the integral of a divergence is the integral of the flux outside the surface [18, 21]. However, as it is explained later in section 2.3.2, since a zero padding is added around the image, then the flux is 0 at every point of the boundaries of the surface. Therefore, equation (12) is the remaining term of equation (11), where the value of $A$ is substituted by its definition in equation (8).

$$F = -\int_{\mathbb{R}^n} \left[ \nabla \cdot \left[\boldsymbol{E}_p - \nabla \left((\nabla \cdot \boldsymbol{E}_p) * V_{mono}\right)\right] \right] U \, d\mu \quad (12)$$

Then, equation (13) distributes de derivative operators according to the properties of the sum and the convolutions.

$$F = -\int_{\mathbb{R}^n} \left[\nabla \cdot \boldsymbol{E}_\mathrm{p} - \nabla^2\left((\nabla \cdot \boldsymbol{E}_\mathrm{p}) * V_\mathrm{mono}\right)\right] U \, d\mu$$
$$F = -\int_{\mathbb{R}^n} \left[\nabla \cdot \boldsymbol{E}_\mathrm{p} - \left((\nabla \cdot \boldsymbol{E}_\mathrm{p}) * \nabla^2 V_\mathrm{mono}\right)\right] U \, d\mu \quad (13)$$

Finally, since $V_\mathrm{mono}$ is the Green's function of $\nabla^2$, then by definition $\nabla^2 V_\mathrm{mono}$ is a Dirac's delta $\delta$ [18]. Knowing that for any function $f$ convoluted with a Dirac's delta $\delta$, we have $f * \delta = f$ [18], equation (14) gives us the final result $F = 0$. Hence, $\boldsymbol{E}_p - \boldsymbol{E}_c$ is orthogonal to any other field. According to Hilbert's theorem, the conservative field $\boldsymbol{E}_c$ has the least error when compared to the perturbed field $\boldsymbol{E}_p$.

$$F = -\int_{\mathbb{R}^n} \left[\nabla \cdot \boldsymbol{E}_\mathrm{p} - \left((\nabla \cdot \boldsymbol{E}_\mathrm{p}) * \delta\right)\right] U \, d\mu$$
$$F = -\int_{\mathbb{R}^n} [\nabla \cdot \boldsymbol{E}_\mathrm{p} - \nabla \cdot \boldsymbol{E}_\mathrm{p}] U \, d\mu \quad (14)$$
$$F = 0$$

This completes the proof that the GFC method allows computing the field $\boldsymbol{E}_c$ and the potential $I_c$ which are the optimal conservative approximation for any perturbed field $\boldsymbol{E}_p$. Hence, the GFC method will always converge to the least-error possible solution, meaning that it is robust to any change or perturbation added to the field. This proof is also valid in the case of an n-dimension image or signal, not just in 2D.

Although we prove that the proposed GFC method is a least-error solver, it does not mean that the cited competing methods are not also least-error solvers. However, section 3.1.4 demonstrates empirically that GFC has consistently lower error than the competing Perez [2] and Tanaka [3] methods, thus supporting the proof that the GFC solver is optimal in the case of added perturbation.

## 2.3 Numerical implementation

The mathematical proof of section 2.2 demonstrated that the proposed GFC method gives the optimal result without any iterative computation, even when a perturbation is added to the gradient. The current section will show how to implement the optimal GFC solver numerically using fast Fourier transforms (FFT).

### 2.3.1  Problems with the Green's function

Although the $n^{th}$ dimension Green's function is defined in equation (5), it cannot be directly applied to an image. The reason is that the function is defined in a continuous infinite space, while images are a bounded discrete space.

Other works propose to use boundary conditions [3, 15] or to find the analytical Green's function for a rectangular space [13]. In our work, we propose using a purely numerical solution, that can also be generalized to non-Laplacian operators.

Advantages of our numerical method are that it is simple to implement, fast to compute and considers the grid structure of the space and the grid nature of the FFT.

### 2.3.2  The numerical Green's function

This subsection shows how to build the numerical Green's function using the convolution theorem and the numerical Fourier transform.

First, the images, gradient, and Laplacian are defined as 2D matrices with an intensity value at each point. For the gradient, there are 2 matrices, one for the horizontal direction and one for the vertical direction. For each pixel in an image, there is an associated Laplacian and gradient.

The numerical gradient and Laplace operators are defined as smaller kernel matrices, which are applied on images via convolution. The numerical Laplace operator is given by equation (15) [7, 8].

$$K_{\nabla^2} = \begin{bmatrix} 0 & -1 & 0 \\ -1 & 4 & -1 \\ 0 & -1 & 0 \end{bmatrix} \quad (15)$$

We also know that, by definition, the Green's function $V_{mono}$ convoluted by the Laplacian operator $K_{\nabla^2}$ should give the Dirac's delta $\delta$ [18]. This relation is shown in equation (16) where $*$ is the convolution operator.

$$(K_{\nabla^2} * V_{mono}) = \delta \qquad (16)$$

We also know that the convolution is defined as the product in the Fourier domain as given in equation (17) [18], where $\mathcal{F}$ is the Fourier transform, $\mathcal{F}^{-1}$ is the inverse Fourier transform, and $A, B$ are any function.

$$A * B = \mathrm{F}^{-1}(\mathrm{F}(A) \circ \mathrm{F}(B)) \qquad (17)$$

Numerically, the Fourier transform is fast and easy to compute using Fast Fourier Transform (FFT) algorithms.

Using equation (16) with the convolution theorem (17), we obtain equation (18). Then we isolate $V_{mono}$ in equation (19) to obtain a mathematical definition of the Green's function $V_{mono}$ in the Fourier domain, which we note $V_{\mathrm{mono}}^{\mathcal{F}}$.

$$\mathcal{F}^{-1}(\mathcal{F}(K_{\nabla^2}) \circ \mathcal{F}(V_{\mathrm{mono}})) = \delta \qquad (18)$$

$$V_{mono}^{\mathcal{F}} \equiv \mathcal{F}(V_{\mathrm{mono}}) = \frac{\mathcal{F}(\delta)}{\mathcal{F}(K_{\nabla^2})} \qquad (19)$$

For this definition to work in a discrete environment, we need the matrices to all be the same size as the image $I$. Hence, we define the zero-padded matrices $\check{K}_{\nabla^2}$ and $\check{\delta}$ in equations (20) and (21), where the top left corner are the $3 \times 3$ Laplacian and Dirac kernels and the rest of the matrices is 0-valued.

$$\check{K}_{\nabla^2} \equiv \begin{bmatrix} 0 & -1 & 0 & \cdots & 0 \\ -1 & 4 & -1 & & \\ 0 & -1 & 0 & & \\ \vdots & & & \ddots & \\ 0 & & & & 0 \end{bmatrix}_{size(I)} \qquad (20)$$

$$\check{\delta} \equiv \begin{bmatrix} 0 & 0 & 0 & \cdots & 0 \\ 0 & 1 & 0 & & \\ 0 & 0 & 0 & & \\ \vdots & & & \ddots & \\ 0 & & & & 0 \end{bmatrix}_{size(I)} \qquad (21)$$

Using the definitions (20) and (21) alongside equation (19), we find the Green's function in the Fourier domain $\check{V}_{\mathrm{mono}}^{\mathcal{F}}$ in equation (22).

$$\check{V}_{mono}^{\mathcal{F}} = \frac{\mathcal{F}(\check{\delta})}{\mathcal{F}(\check{K}_{\nabla^2})} \qquad (22)$$

Finally, using the Laplacian solver of equation (4) we can solve the least-error potential $I_c$ from its Laplacian $L$. The result is given in equation (23), where $\mathcal{R}$ is the real part of a complex number and $\circ$ is the Hadamard element-wise product. We note that $I_c = I$ if the right integration constant $c$ is used.

$$I_c = \mathcal{R}\left(\mathcal{F}^{-1}(\mathcal{F}(L) \circ \check{V}_{mono}^{\mathcal{F}})\right) + c \qquad (23)$$

In the cases where the boundaries need to be preserved, then it is suggested to add 3-pixel padding to $I$ before passing to the gradient domain, then retrieve the constant $c$ such as the padded region in $I_c$ has a value of 0.

We validated numerically equation (23) by computing the Laplacian $L$ of the 1000 images from the ECSSD dataset [22], then computing $I_c$ using the Green's function $\check{V}_{\mathrm{mono}}^{\mathcal{F}}$. We found the root mean square error (RMSE) to be 0.011 on 256 levels, which is 0.004% of numerical error, which is negligible.

### 2.3.3 A universal convolution reversal?

At first sight, the equations developed in the previous section seems to reverse any convolution kernel $K$, since equation (19) finds the reverse kernel of any operator. However, the Green's function is only defined for differential operators, meaning that non-differential operators do not necessarily have a reverse.

For example, reversing the popular Sobel gradient operator [8, 23] can be done with equation (19), but there will be a significant error on the regions of high gradient. This is because the Sobel operator is a blurred version of the gradient operator, which dissipates high frequencies and cannot be reversed completely. Hence, the resulting image $I_c$ from equation (23) for a Sobel gradient is a blurred version of $I$.

### 2.3.4 Computation complexity

As shown previously, the Laplacian $L$ is solved using equation (23), where the only operations consist of the Fourier transforms $\mathcal{F}$ and the element-wise product $\circ$. In this case, the computation complexity will be dominated by the FFT algorithms with a computation complexity of $O(n \log n)$ [23], where $n$ is the total number of pixels.

Although the complexity is not linear, the logarithmic term becomes less important when the number of pixels is near the million, which is typical for images.

## 3 Applications in computer vision

There are many already-proven applications of the Laplacian solvers in computer vision, including seamless cloning, seamless composite, and animated diffusion curves [1–3], etc. Those applications are part of a branch called gradient-domain image editing (GDIE) [12].

Since they are already demonstrated, we will only focus mainly on showing the proof-of-concept of the GFC with some comparison to Perez [2], Jeschke [1] and Tanaka [3] methods. Using the development of the previous section, we know that equation (23) is a least-error solver of the Laplacian. We will also demonstrate that the proposed approach is significantly faster than competing methods and that it can be leveraged for machine learning (ML) applications.

### 3.1 Solving the image Laplacian

In this section, we summarize the GDIE process using our proposed GFC method. Then, we benchmark the solver computation time and error against competing methods.

#### 3.1.1 GDIE process summary

**Fig. 1** shows a summary of the process used to solve the modified gradient for GDIE applications. All those steps are simple to implement in OpenCV and Matlab since they mostly use already available functions in their respective computer vision toolboxes. Some of the process summary steps, such as the gradient editing and color correction, will be discussed in later sections.

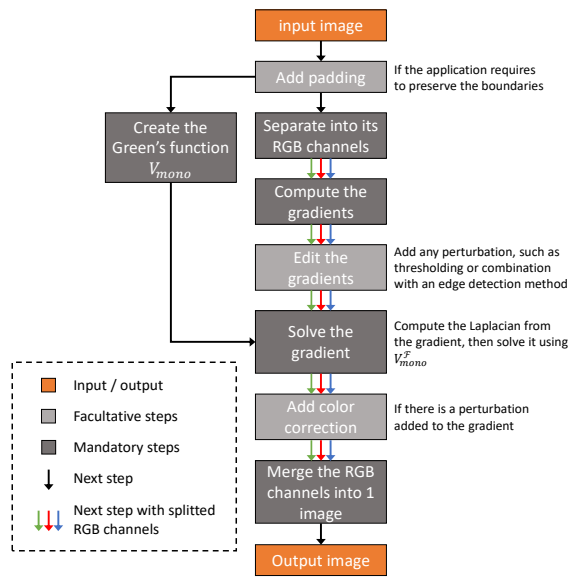

**Fig. 1**  Process summary of the gradient domain image editing.

#### 3.1.2 Pseudo-code

In this section, we demonstrate the simplicity of GFC by providing some Python-based pseudo-codes.

First, Algorithm A shows how to compute the Green's function `green_F` for an image of size `image_size` using equation (19). Then, Algorithm

B shows how the previous Green's function is used to solve the given Laplacian `padded_L` with equation (23).

For both algorithms, we must keep in mind that the 2D FFT `fft` and inverse FFT `ifft` produce complex outputs, meaning that the products and division must be used accordingly.

The code simplicity allowed us to implement the solver using Matlab, C++ (OpenCV) and Python (Tensorflow and Pytorch).

---

**Algorithm A.** Python based pseudo-code for computing Green's function

```
# Inputs:
# image_size: The size of the image
# pad: The padding to add around the image

# Find the size of the desired matrices
pad ← 4
green_function_size ← image_size + 2 * pad

# Create the Dirac and Laplace kernels
dirac ← zeros(green_function_size)
dirac[1, 1] ← 1
laplace ← zeros(green_function_size)
laplace[0:3, 0:3] ← [[0,  1,  0],
                    [1, -4,  1],
                    [0,  1,  0]]

# Compute the Green's function
green_F ← fft(dirac) / fft(laplace)
```

---

**Algorithm B.** Python based pseudo-code for solving the padded Laplacian

```
# Inputs:
# padded_L: The Laplacian of the padded image
# green_F: The result of Algorithm A

# Solving the padded Laplacian
I ← ifft(fft(padded_L) * green_F)

# Integration constant and unpadding
I ← I - I[0, 0]
I ← I[pad:-pad, pad:-pad]
```

---

### 3.1.3 Computation time benchmark

As explained previously, the computation time of our proposed GFC method is low since FFT is highly optimized on CPUs and GPUs [8, 9]. For example, the computation time is around 18ms on MATLAB® with an *Intel® i7-6700K* processor for a gray image (single channel) of resolution of 801x1200. Also, using MATLAB's GpuArray with the GPU *nvidia® GTX 1080 Ti*, the computation time when the overhead is eliminated is around 0.8ms.

In all the implementations, we used 32-bit floating points, since a double precision is not required.

**Our method (GFC).** In **Fig. 2**, the total time for the GFC is noted 1.3ms, which includes 0.5ms for the preparation such as verifying the parameters and sending the matrices to the GPU. The remaining 0.8ms is used for solving the gradient.

For the GFC method, the computation time in **Fig. 2** does not include the computation of the optimal Green's function $\check{V}_{\text{mono}}^{\mathcal{F}}$ since it can be pre-computed with equation (22). The time to build it is 5ms on the GPU and 36ms on the CPU. Even if $\check{V}_{\text{mono}}^{\mathcal{F}}$ is not pre-computed, the method is still fast enough to out-perform any competing algorithm, since $\check{V}_{\text{mono}}^{\mathcal{F}}$ is computed only once for the 3 channels of an image.

With the logarithmic scale of **Fig. 2**, we can observe that the proposed GFC method is orders of magnitude faster than competing algorithms, such as Perez et al. [2] or Jeschke et al. [1].

**Our method (GFC) using Pytorch batches.** Since one of our objectives is to develop a method compatible with CNN, we decided to implement our method on the Pytorch [24] machine learning library. For the Pytorch implementation, we use batches of 100 different images of size 801x1200, since it is similar to how CNN use batches of features and can be useful for video editing. On a CPU, we found that the batches did not improve the computation time. However, on a GPU, the computation time for a single image (~0.9ms) was almost identical to the batch of 100 images (~1.0ms). Hence, the average time per image is 0.01ms as noted in **Fig. 2**. Since 100 images are near the memory limits of our GPU, the 0.01ms per image is the fastest we can achieve in parallel.

**Perez method.** The Perez [2] algorithm is downloaded from MathWorks [25], and later optimized to use the full capacities of MATLAB, but the matrix inversion alone takes 1770ms with another 1270ms to build the sparse matrix. It has no GPU implementation.

**Tanaka method.** The Tanaka [3] algorithm is written by the author and is downloaded from MathWorks [26]. The computation time on the CPU to perform the cosine-transforms required to solve the Laplacian is around 292ms with a preparation time of 2ms. Furthermore, an additional time of 85ms is added to compute the cosine-transform solver, but it is not included in **Fig. 2** since it can be pre-computed.

**McCann and Pollard multigrid method.** Their method could not serve as a benchmark in its current form. Indeed, the provided code is implemented on older hardware and 32-bit libraries, and since their binaries only implement diffusion curves. According to their work [11], their proposed multi-grid solver requires around 10 iterations to converge, so that the process lasts ~110ms for an $801 \times 1200$ image on the older GPU *nvidia® GeForce 8600 GTS* with a performance of 93 GFLOPS [27]. On the *nvidia® GTX 1080 Ti* with 11340 GFLOPS [27], the fastest expected time is 0.9ms $\left(\frac{110 \cdot 93}{11340}\right)$. This computation time is only possible if all the CUDA cores are used since the GPU clock is only twice the speed [27]. Hence, the method proposed in this paper, which solves the gradient in $0.8\,\text{ms}$ (without preparation time), is expected to be equal or faster than McCann for a single image. Furthermore, the proposed method is parallelizable to 100 images in 1 ms, but we do not know if the same is true for McCann.

**Jeschke method (diffusion curve only).** The Jeschke algorithm is provided with their paper [1], but it is only implemented for diffusion curves. Hence, an ideal comparison with their algorithm is not possible and the time is not included in **Fig. 2**. Their algorithm was benchmarked to 6.2 ms on GPU and $476.2\,\text{ms}$ on CPU for a single channel computation. Although the comparison is not ideal, this is orders of magnitude slower than our implementations.

**Green's function based methods.** The methods proposed by Sun et al. [13, 14] and Ilbery et al. [15] are both based on Green's function diffusion. However, they are only implemented for diffusion curves and cannot directly work with discrete grids for image editing purposes. This is because they compute the Green's function in a continuous bounded 2D space for application on vector curves. Therefore, a comparison is not directly possible.

**Other methods.** Other methods such as the one proposed by Bhat et al. do not perform real-time image editing as stated in their paper [12], which means that it is definitely slower than the proposed approach.

In summary, the proposed GFC algorithm runs orders of magnitude faster for discrete images than competing algorithms. Compared to the Tanaka method, the improvement is 16x faster on CPU and 172x faster on GPU. Furthermore, our GPU Pytorch implementation shows that the computation time for batches of 100 images is the same as for a single image. Since the method is fast, we expect that a major part of the running time in a real application will be due to overheads and verifications.

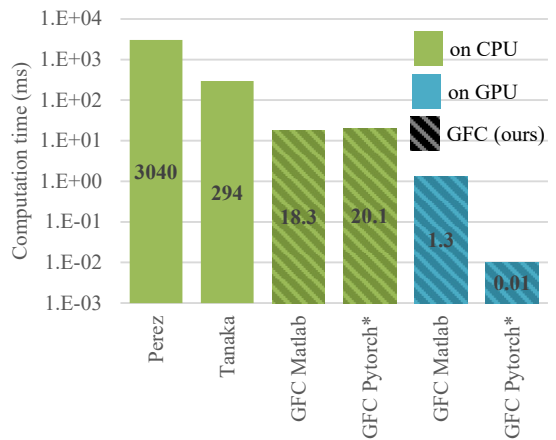

**Fig. 2** Computation time (ms) in logarithmic scale for a single channel gradient solving of resolution of 801x1200, including the preparation time. The Perez [2] and Tanaka [3] methods have no GPU implementation. The Pytorch* implementation is tested on batches of 100 images, and the total time is divided by 100.

### 3.1.4 Non-conservative solver benchmark

We proved in section 2.2 that the Green's function is the least-error solver for any non-conservative fields $E_p$. In this section, we demonstrate empirically that our method has less error than the Perez [2] and Tanaka [26] methods.

To demonstrate it, we use the 1000 images from the ECSSD dataset [22] and compute the gradients $E$. We modify the gradients by setting any value below a given threshold to 0, with thresholds at 10%, 30% and 50%, resulting in $E_p$. Then, we solve $E_p$ using the different methods and find the new gradient $E_c$. Finally, we compute the root mean squared error (RMSE) between the gradients using equation (24).

We observe on **Fig. 3** that the RMSE of our GFC method is consistently lower than competing methods. For the 10% threshold, GFC has an RMSE 16% lower than Tanaka and 76% lower than Perez.

$$\text{RMSE} = \text{mean}\left((E_c - E_p)^2\right) \qquad (24)$$

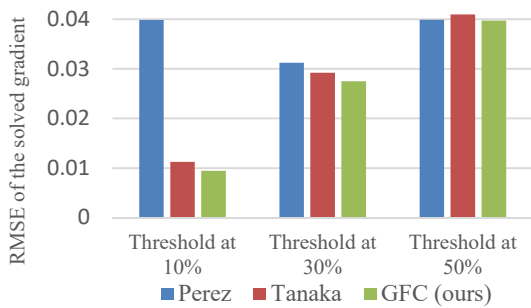

**Fig. 3** Comparison of the RMSE between $E_c$ and $E_p$, where the thresholds are the perturbation used to generate $E_p$. The displayed values are the mean of the RMSE on the 1000 images of the ECSSD dataset [22].

### 3.2 Gradient-domain image editing

From the mathematical proof presented in section 2.2, the GFC proved to be the least-error solver for any perturbed gradient. In the case of GDIE, the gradient perturbation is voluntary. It is mainly used for applications such as Poisson blending, diffusion curves [1–3] and edge editing [12].

This section will show the performance of the method for Poisson blending, as well as additional possible gradient-domain applications such as the proposed gradient domain merging (GDM) based on the work of Bhat et al. [12]. Those applications can potentially be used in image/video editing software, as well as image pre-processing.

#### 3.2.1 Poisson blending

Poisson blending is a type of GDIE that allows merging the gradient of 2 different images, such that the blending is seamless. Since the proposed GFC approach has a low computation time for large images, as demonstrated in section 3.1.3, our implementation of the Poisson blending uses a blend region that is bounded by the total size of the image. This means that if the cropping region passes through a high gradient region, our method is better at compensating the error. This is shown inside the blue circle of **Fig. 4** where the GFC approach solves smoothly the cropped edge. Also, the GFC blending appears more natural since the left side of the stamp is more transparent.

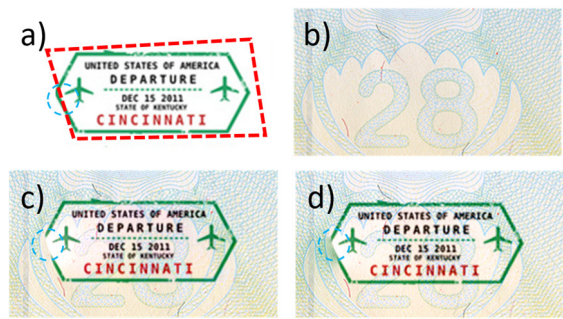

**Fig. 4** Example of Poisson Blending application; (a) Stamp to copy, with the red-dotted lines being the cropping region and the blue dotted circle being a region of the stamp that is accidentally cropped; (b) destination image; (c) Poisson blending from Perez algorithm [2, 25]; (d) Proposed GFC blending.

In other cases where the cropping region does not pass through a high gradient, the results of the proposed GFC method is identical to the Perez method.

#### 3.2.2 Preserving the coloration

The equation (23) presented an optimal Laplacian solver in the Fourier domain. In comparison, the literature proposes mostly iterative methods on the Laplacian [1–3], which gives an advantage for our method by making it faster and easier to implement. However, computing the Laplacian from the perturbed gradient requires an additional computing step to preserve the brightness and contrast.

The problem when editing the gradient in an image is that the desired potential is not necessarily the result given by $V_E$ since we want to preserve the color information. Hence, we define a new corrected potential $I_{c,corr}$ in equation (25), where $\sigma$ indicates the standard deviation and the top bar "¯" indicates the average. As stated in section 2.3.2, it is possible to add any constant to $I_c$ without changing the validity of the equation, which means that the addition and subtraction of equation (25) do not affect the potential. For the $\sigma$ ratio, it is meant to preserve the initial contrast of the image, and it simply changes the norm of the gradient by a constant factor.

$$I_{c,corr} = (I_c - \bar{I_c})\frac{\sigma(I)}{\sigma(I_c)} + \bar{I} \qquad (25)$$

In case no perturbation is added to the gradient, we have $I_c \approx I$. This means that almost no correction will be added to $I_c$, resulting in $I_{c,corr} \approx I_c$.

### 3.2.3 Gradient thresholding

Using the color preservation of equation (25), **Fig. 5** shows an of thresholding the gradient at 10% of the highest possible gradient and computing the solving the new image with equations (23) and (25). We can see that most features of the initial image are preserved, but that there are less texture and fine elements.

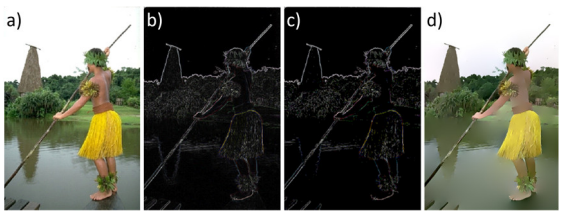

**Fig. 5** Example of gradient thresholding and solver steps. (a) Original image; (b) Gradient $|\mathbf{E}|$; (c) Thresholded gradient at 10%; (d) Solved image $I_{c,corr}$.

**Fig. 6** shows 2 more examples of GDIE with a 10% gradient thresholding. In those images, the castle reflection is completely erased, along with the clouds. For the leopard picture, almost all the background information is erased except for the leopard.

The differences between our proposed Laplacian solver GFC and the one proposed by Perez [2] are negligible in the case of gradient removal. Hence, we do not present comparison images since the differences are imperceptible to the human eye.

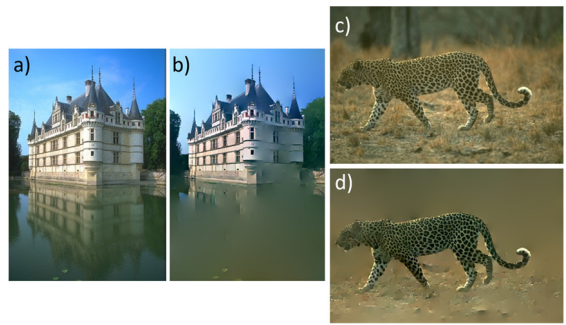

**Fig. 6** Examples of solved images $I_{c,corr}$ after the gradient threshold at 10%. (a) Image of a castle; (b) Solved castle image after 10% gradient threshold; (c) Image of a leopard; (d) Solved leopard image after 10% gradient threshold.

### 3.2.4 Gradient domain merging (GDM)

In this section, we present a method of editing an image by merging edges information with gradient information.
A similar approach was used by Bhat et al. [12], which computed the salient gradient map to enhance the original gradient via cosine similarity. What we propose instead is to use the edges produced by ML algorithms and merge them to the gradient via a geometric average.

The motivation of using machine learning edges prediction is that we believe future work could benefit from implementing the Green's function inside ML algorithms. A simple example would be to enhance the contrast of the important objects via GDM, thus making it easier for the ML method to detect the object.

The proposed GDM approach consists of combining the gradient with the edge information using a weighted geometric average defined in (26).

The product enhances the gradient where edges are present but reduces them where edges are not present. In the equation, $E_p$ is the perturbed gradient, $E$ is the original gradient, $C$ is the intensity of the edge detection, ∘ is the element-wise product, and $\alpha = [0,1]$ is the weight associated to the geometric mean. Also, the orientation of the perturbed gradient $\theta_{E_p}$ is equal to the orientation of the original gradient $\theta_E$. A higher $\alpha$ attributes more weight to the edges, while a lower $\alpha$ attributes more weight to the gradient.

$$|E_p| = |E|^{1-\alpha} \circ C^\alpha$$
$$\theta_{E_p} = \theta_E \tag{26}$$

#### 3.2.4.1 Contrast enhancement between objects

**Fig. 7** shows the different steps involved in the GDM method. The perturbed field $E_p$ is produced by equation (26) and solved by equation (25). We observe that the GDM produces a loss of texture, but that the contrast between the objects are enhanced.

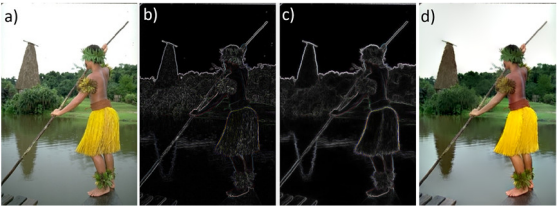

**Fig. 7** Examples of the steps involving the GDM equation (26). (a) Original image; (b) $|E|$: Gradient; (c) $|E_p|$: Gradient merged with the SE method [16] and $\alpha = 0.5$; (d) Solved image.

In **Fig. 8**, we can see the effect of using different $\alpha$ parameters. The higher the parameter $\alpha$ is chosen, the stronger is the contrast between objects. However, a higher $\alpha$ creates discoloration in the image. This is because a higher $\alpha$ produces a field that is too different from the original field, which yields in undesired coloring and brightness artefacts.

In **Fig. 9**, we can observe more examples of GDM using $\alpha = 0.5$. We see that the deep learning edges RCF [17] produces higher contrast than the between objects than the random forest SE edges [16].

In **Fig. 10**, we can observe that Bhat [12] method of saliency sharpening enhances the folds of the clothing and the lines in the background. This is opposite to our method which reduces the folds and the background texture but enhances the colors of the foreground objects. This demonstrates that our method is fundamentally different than previously proposed edge enhancement methods and should not be used for the same purposes.

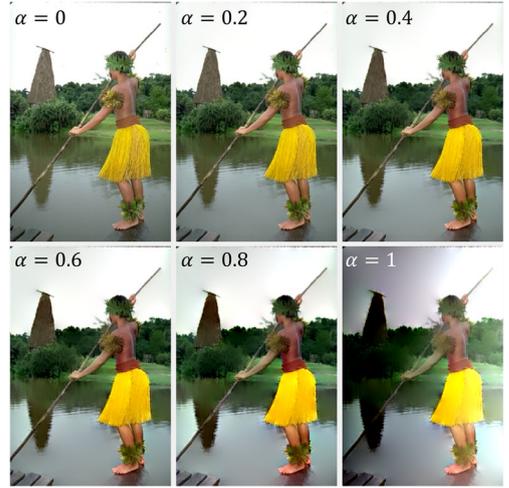

**Fig. 8** Example of GDM using equation (26) with a random forest edge detector [16] and varying parameter $\alpha$.

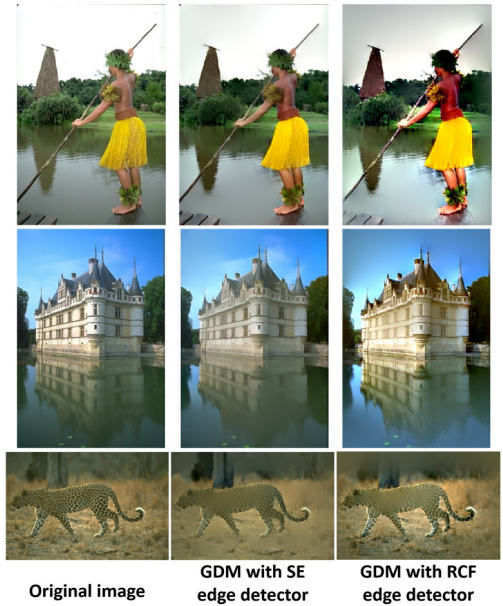

**Fig. 9** Examples of solved images $I_{c,corr}$ with a perturbed gradient from equation (26) with edges information from SE method [16] and RCF method [17] and $\alpha = 0.5$.

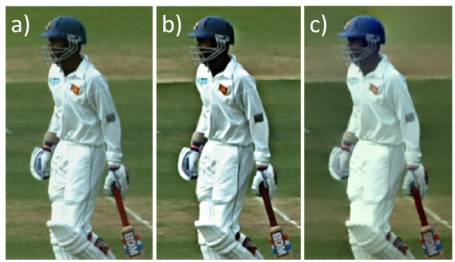

**Fig. 10** Comparison of edge merging methods. (a) Original image; (b) Saliency sharpening from Bhat et al. [12]; (c) Enhancement using our GDM method with $\alpha = 0.5$ and SE edge detector.

#### 3.2.4.2  Painting effect using thin edges

For the GDM method, it is also possible to thin the edges before merging them with the gradient. This thinning is often called non-maximal suppression (NMS) and is natively implemented in some edge detection such as SE [16]. Applying NMS to the edges removes almost completely the texture information, meaning that the solved image resembles a painting, as observed in **Fig. 11**.

Since the thin edges $C$ do not necessarily intersect the gradient, we thicken the gradient $E$ by using a Gaussian filter with a standard deviation $\sigma = 1$.

In **Fig. 11**, we compare our GFC method and the one proposed by Perez [2, 25]. First, we notice that the GFC approach has better color preservation than the Perez method. For example, we observe on the person image that the sky has a gradient of different colors. We also observe that the castle image has many small coloration artifacts inside the castle and at the top of the sky. For the leopard image, both methods yield similar results. Thus, the proposed GFC method produces a more natural painting effect since it is more accurate on fine details and color restoration than the competing Perez algorithm [2].

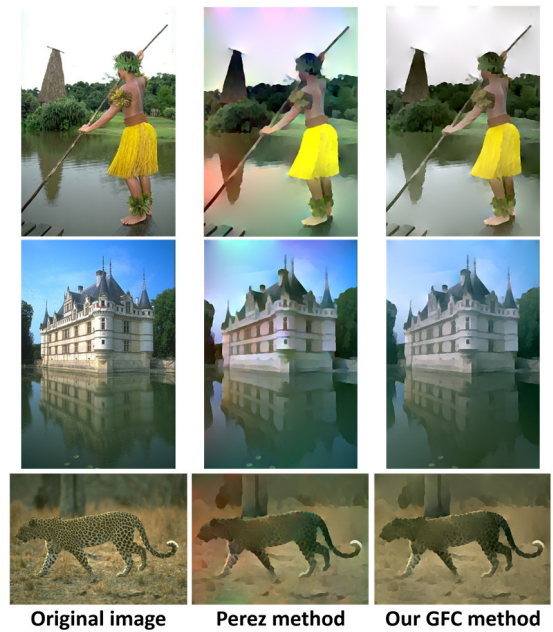

**Fig. 11** Examples of solved images $I_{c,corr}$ with a perturbed gradient from equation (26) with edges information from SE method [16] with NMS and $\alpha = 0.5$.

## 4 Future work

In this section, we briefly discuss possible future work of our research concerning the convolutional neural network (CNN) applications and tensor processing unit (TPU) implementation.

**Machine learning applications.** As shown in **Fig. 2**, one of the advantages of our method is that the computation time is orders of magnitude faster than competing methods, and 100 folds parallelizable. Furthermore, Algorithm B shows that the code is very simple to implement if a 2D FFT is available. These advantages will allow future work to use the GFC inside CNN, thus allowing the networks to natively learn gradient-domain image editing. In fact, our Pytorch and Tensorflow implementations of GFC can be easily integrated inside a network since the backward propagation of the FFT is natively available.

**TPU implementation.** To further improve the computation speed and the parallel processing of our method, it will be interesting to implement it on TPU. These new processors allow parallelizing more operations by using 16 bits floating points, which can heavily benefit the computation of FFT [28].

## 5 Conclusion

This study detailed the development of the GFC method, which allows solving any field or Laplacian for gradient domain image editing purposes. First, we explained the theory behind the Green function convolution (GFC), and we mathematically proved in section 2.2 that it is the least error solver. Then, we demonstrated empirically on 1000 images that the RMSE error is negligible with a value of 0.004%. Moreover, **Fig. 3** also showed that the solver error on non-conservative fields is consistently lower than competing methods. **Fig. 2** also showed that the method is almost instantaneous with a computation time of 1ms for the parallel processing of 100 images with resolution 1200x801. Finally, we demonstrated different use-cases of gradient domain image editing and introduced GDM, the first method of merging learned edges with gradients for texture removal and contrast enhancement.

In summary, this study allowed to build a robust and fast way to edit an image from its gradient which can be used in many applications. The code is fast enough to have a negligible impact on the computation time and simple enough to be implemented in any language. Future works could focus on more concrete applications, such as supervised image/video editing and machine learning applications.

## Acknowledgment

The authors are grateful to NSERC, through the discovery grant program RGPIN-2014-06289, and FRQNT/INTER for their financial support.

## References

1. Jeschke, S., Cline, D., Wonka, P.: A GPU Laplacian Solver for Diffusion Curves and Poisson Image Editing. In: ACM SIGGRAPH Asia 2009 Papers. pp. 116:1–116:8. ACM, New York, NY, USA (2009). https://doi.org/10.1145/1661412.1618462.
2. Pérez, P., Gangnet, M., Blake, A.: Poisson Image Editing. In: ACM SIGGRAPH 2003 Papers. pp. 313–318. ACM, New York, NY, USA (2003). https://doi.org/10.1145/1201775.882269.
3. Tanaka, M., Kamio, R., Okutomi, M.: Seamless Image Cloning by a Closed Form Solution of a Modified Poisson Problem. In: SIGGRAPH Asia 2012 Posters. pp. 15:1–15:1. ACM, New York, NY, USA (2012). https://doi.org/10.1145/2407156.2407173.
4. Beaini, D., Achiche, S., Cio, Y.-S.L.-K., Raison, M.: Novel Convolution Kernels for Computer Vision and Shape Analysis based on Electromagnetism. ArXiv180607996 Cs. (2018).
5. Beaini, D., Achiche, S., Nonez, F., Raison, M.: Computing the Spatial Probability of Inclusion inside Partial Contours for Computer Vision Applications. ArXiv180601339 Cs Math. (2018).
6. Dominique Beaini, Sofiane Achiche, Maxime Raison: Object analysis in images using electric potentials and electric fields.
7. Forsyth, D., Ponce, J.: Computer Vision: A Modern Approach. Pearson (2012).
8. Laganière, R.: OpenCV 2 Computer Vision Application Programming Cookbook. Packt Publishing, Birmingham (2011).
9. Corke, P.: Robotics, Vision and Control: Fundamental Algorithms in MATLAB. Springer Science & Business Media (2011).

10. Orzan, A., Bousseau, A., Barla, P., Winnemöller, H., Thollot, J., Salesin, D.: Diffusion Curves: A Vector Representation for Smooth-shaded Images. Commun ACM. 56, 101–108 (2013). https://doi.org/10.1145/2483852.2483873.
11. McCann, J., Pollard, N.S.: Real-time Gradient-domain Painting. In: ACM SIGGRAPH 2008 Papers. pp. 93:1–93:7. ACM, New York, NY, USA (2008). https://doi.org/10.1145/1399504.1360692.
12. Bhat, P., Zitnick, C.L., Cohen, M., Curless, B.: GradientShop: A Gradient-domain Optimization Framework for Image and Video Filtering. ACM Trans Graph. 29, 10:1–10:14 (2010). https://doi.org/10.1145/1731047.1731048.
13. Sun, T., Thamjaroenporn, P., Zheng, C.: Fast Multipole Representation of Diffusion Curves and Points. ACM Trans Graph. 33, 53:1–53:12 (2014). https://doi.org/10.1145/2601097.2601187.
14. Sun, X., Xie, G., Dong, Y., Lin, S., Xu, W., Wang, W., Tong, X., Guo, B.: Diffusion Curve Textures for Resolution Independent Texture Mapping. ACM Trans Graph. 31, 74:1–74:9 (2012). https://doi.org/10.1145/2185520.2185570.
15. Ilbery, P., Kendall, L., Concolato, C., McCosker, M.: Biharmonic Diffusion Curve Images from Boundary Elements. ACM Trans Graph. 32, 219:1–219:12 (2013). https://doi.org/10.1145/2508363.2508426.
16. Dollár, P., Zitnick, C.L.: Fast Edge Detection Using Structured Forests. IEEE Trans. Pattern Anal. Mach. Intell. 37, 1558–1570 (2015). https://doi.org/10.1109/TPAMI.2014.2377715.
17. Cheng, M.M.: Richer Convolutional Features for Edge Detection, https://mmcheng.net/rcfedge/, (2017).
18. Arfken, G.B., Weber, H.J.: Mathematical Methods for Physicists, 6th Edition. Academic Press, Boston (2005).
19. Canny, J.: A Computational Approach to Edge Detection. IEEE Trans. Pattern Anal. Mach. Intell. PAMI-8, 679–698 (1986). https://doi.org/10.1109/TPAMI.1986.4767851.
20. Weisstein, E.W.: Projection Theorem, http://mathworld.wolfram.com/ProjectionTheorem.html.
21. Feynman, R.P., B, F.R.P.S.M.L.L.R., Leighton, R.B., Sands, M.: The Feynman Lectures on Physics, Desktop Edition Volume II: The New Millennium Edition. Basic Books (2013).
22. Shi, J., Yan, Q., Xu, L., Jia, J.: Hierarchical Image Saliency Detection on Extended CSSD. IEEE Trans. Pattern Anal. Mach. Intell. 38, 717–729 (2016). https://doi.org/10.1109/TPAMI.2015.2465960.
23. Szeliski, R.: Computer Vision: Algorithms and Applications. Springer Science & Business Media (2010).
24. PyTorch, https://www.pytorch.org.
25. Afifi, M.: Poisson image editing - File Exchange - MATLAB Central, https://www.mathworks.com/matlabcentral/fileexchange/62287.
26. Tanaka, M.: Fast seamless image cloning by modified Poisson equation - File Exchange - MATLAB Central, https://www.mathworks.com/matlabcentral/fileexchange/39438.
27. TechPowerUp, https://www.techpowerup.com/gpu-specs/.
28. Sorna, A., Cheng, X., D'Azevedo, E., Won, K., Tomov, S.: Optimizing the Fast Fourier Transform Using Mixed Precision on Tensor Core Hardware. In: 2018 IEEE 25th International Conference on High Performance Computing Workshops (HiPCW). pp. 3–7 (2018). https://doi.org/10.1109/HiPCW.2018.8634417.